\documentclass[letterpaper, 10 pt, conference]{ieeeconf}  

\usepackage{hyperref}
\hypersetup{hidelinks,
	colorlinks=true,
	allcolors=black,
	pdfstartview=Fit,
	breaklinks=true}

\usepackage{amsmath}
\usepackage{booktabs}
\usepackage{multirow}
\usepackage{graphicx}
\usepackage{amsfonts,amssymb}
\usepackage{romannum}

\IEEEoverridecommandlockouts                              

\usepackage{graphics} 
\usepackage{epsfig} 
\usepackage{mathptmx} 
\usepackage{times} 
\usepackage{amsmath} 
\usepackage{amssymb}  

\usepackage{multirow}
\usepackage[flushleft]{threeparttable}  
\usepackage[table]{xcolor}
\usepackage{xcolor}
\usepackage{gensymb}
\usepackage{siunitx}
\usepackage{float}
\usepackage{tabularx}

\definecolor{Gray}{gray}{0.9}
\newcommand{\grayrule}{\arrayrulecolor{black!30}\midrule\arrayrulecolor{black}}

\title{\LARGE \bf
What Really Matters for Learning-based LiDAR-Camera Calibration
}

\author{Shujuan Huang, Chunyu Lin$^{\dagger}$, Yao Zhao 
\thanks{This work was supported by  the National Natural Science Foundation of China (Nos. 62172032,  62120106009).}
\thanks{Institute of Information Science, Beijing Jiaotong University, Beijing 100044, China. Visual Intelligence + X International Joint Laboratory of
the Ministry of Education. Email: \{shujuanhuang, cylin, yzhao\}@bjtu.edu.cn}
\thanks{$^{\dagger}$ Corresponding author.}
}

\begin{document}

\maketitle
\thispagestyle{empty}
\pagestyle{empty}

\begin{abstract}
Calibration is an essential prerequisite for the accurate data fusion of LiDAR and camera sensors. Traditional calibration techniques often require specific targets or suitable scenes to obtain reliable 2D-3D correspondences. To tackle the challenge of target-less and online calibration, deep neural networks have been introduced to solve the problem in a data-driven manner. While previous learning-based methods have achieved impressive performance on specific datasets, they still struggle in complex real-world scenarios. Most existing works focus on improving calibration accuracy but overlook the underlying mechanisms. In this paper, we revisit the development of learning-based LiDAR-Camera calibration and encourage the community to pay more attention to the underlying principles to advance practical applications. We systematically analyze the paradigm of mainstream learning-based methods, and identify the critical limitations of regression-based methods with the widely used data generation pipeline. Our findings reveal that most learning-based methods inadvertently operate as retrieval networks, focusing more on single-modality distributions rather than cross-modality correspondences. We also investigate how the input data format and preprocessing operations impact network performance and summarize the regression clues to inform further improvements.
\end{abstract}

\section{INTRODUCTION}
Multi-modal systems integrating cameras and LiDAR are widely used for their complementary perceptual features. Cameras can obtain high-resolution images with rich texture information, while LiDAR can capture precise but sparse geometric measurements. The combination of LiDAR and cameras is pivotal in robotics \cite{zheng2022fast}, autonomous driving \cite{liu2023bevfusion}, and intelligent transportation infrastructure \cite{mateen2022smart}. Various data fusion strategies of 2D images and 3D point clouds have proven to be effective in numerous tasks, such as object detection \cite{bai2022transfusion}, semantic segmentation \cite{liu2023uniseg}, vehicle localization \cite{wang2022rail}, and occupancy prediction \cite{wang2023openoccupancy}. An important prerequisite for achieving effective multi-modal data fusion is the accurate temporal and spatial calibration.

\begin{figure}
\centering
\includegraphics[width=7cm]{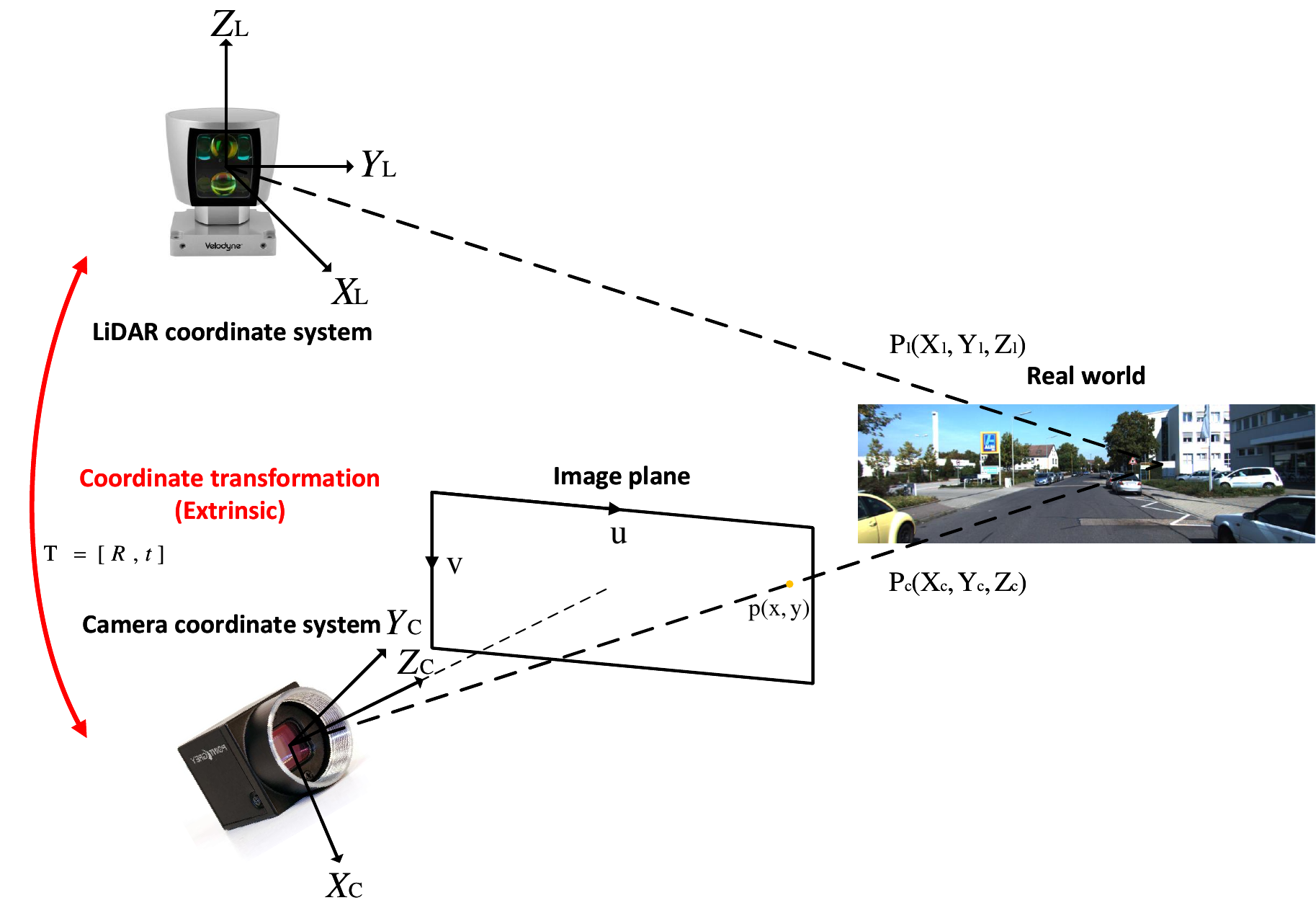}
\caption{The transformation between a point cloud captured from LiDAR and an image captured by camera. The target of LiDAR-Camera calibration is to estimate the coordinate transformation.}
\label{fig1}
\vspace{-0.5cm}
\end{figure}

\begin{figure*}[htbp]
\centering
\includegraphics[width=16cm]{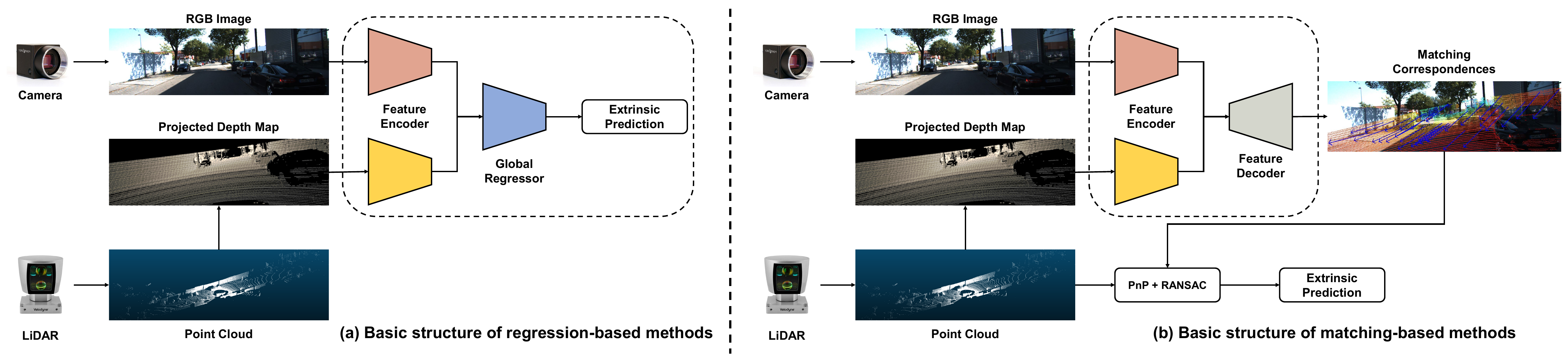}
\caption{Two mainstream learning-based LiDAR-Camera calibration framework, from the perspective of the output format of the network. Regression-based methods predict extrinsic parameters directly in an end-to-end manner. Matching-based methods predict correspondences af first then resolve the extrinsic by explicit geometry solver. (a) The regression-based paradigm (b). The matching-based paradigm.}
\label{fig2}
\vspace{-0.5cm}
\end{figure*}
As temporal and intrinsic calibration can be achieved through joint hardware and software synchronization and well-established methods \cite{zhang2000flexible}, researchers tend to focus on determining extrinsic parameters. For LiDAR-Camera calibration, target-based calibration strategies rely on specific calibration targets with well-designed patterns. With the corresponding 2D pixel locations and 3D point coordinates, extrinsic parameters can be estimated by solving the Perspective-n-Point (PnP) problem, as shown in Fig. \ref{fig1}. However, target-based methods can only be applied offline due to their reliance on calibration targets and often still require manual intervention.

To compensate for the gradual extrinsic deviations caused by shocks or vibrations during daily operation, target-less online calibration methods have been proposed to correct this drift. Intuitive solutions \cite{moghadam2013line,levinson2013automatic} attempt to use common features to replace calibration targets. Other works \cite{nagy2019sfm,ma2021crlf} involve Structure-from-Motion (SfM) or hand-eye calibration techniques to optimize the extrinsic parameters. Although these methods extend the applicability, they often presuppose certain assumptions to provide sufficient geometric constraints. Given the complexity of the real world, it is difficult to guarantee both robustness and accuracy.

Following the development of deep learning techniques, learning-based calibration \cite{schneider2017regnet,iyer2018calibnet,cattaneo2019cmrnet} methods have been proposed to improve calibration accuracy and reduce dependence on specific targets and scenarios. Current learning-based approaches can be divided into two main categories: regression-based and matching-based. Regression-based methods take data from two modalities as input and output extrinsics directly in an end-to-end manner. On the contrary, matching-based methods are more akin to traditional methods and adopt related techniques to obtain dense or sparse correspondences. Despite various modifications designed to enhance performance, we observe that the fundamental principle of how networks learn to estimate the extrinsics has been overlooked.

In this paper, we systematically analyze the paradigms of learning-based methods. We demonstrate that regression-based methods function more like retrieval networks and do not effectively perform cross-modal matching. By adopting a basic framework that removes other auxiliary modules, we set up a series of experiments for verification. Furthermore, we reconsider commonly neglected aspects of previous implementations. We argue that commonly employed data generation methods have critical limitations. To confirm this, we introduce cropping and mixing operations to make the generated dataset more challenging, leading to noticeable performance degradation. Concurrently, we investigate the crucial cues for parameter regression to better understand their working principles. Our key contributions are as follows:
\begin{itemize}
\item We systematically analyze the paradigm of learning-based LiDAR-Camera calibration and elucidate the working principles of regression-based methods.
\item We verify that the typical data generation pipeline has critical limitations and identify crucial clues for regression-based methods.
\item We demonstrate that current regression-based methods exhibit poor generalization due to overfitting to specific datasets and call for more community attention to advance towards real applications.
\end{itemize}

\section{RELATED WORKS}
Generally, there are three main categories of calibration methods: target-based, target-less, and learning-based. To some extent, learning-based methods can be considered as a specialized form of target-less methods. Given that learning-based methods are gradually taking the lead, we categorize them separately as a distinct class.

\subsection{Target-Based Calibration}
Target-based LiDAR-Camera calibration aims to find accurate 2D-3D correspondences utilizing calibration targets that can be simultaneously detected in both modalities. Early stage target-based methods \cite{zhang2004extrinsic,unnikrishnan2005fast,pandey2010extrinsic} were inspired by camera intrinsic calibration techniques using chessboard patterns.Since then, numerous calibration targets based on different patterns have been introduced to improve the accuracy of feature extraction. Typical examples include plane boards with holes \cite{yan2023joint,liu2023robust}, triangles \cite{debattisti2013automated,bu2021calibration}, polygons \cite{liao2018extrinsic}, spheres \cite{zhang2024automatic} and ArUco \cite{dhall2017lidar}. Despite improvements in accuracy, the specialized designs aimed at providing more geometric constraints make the production of targets more challenging. Apart from modifying calibration targets, efforts \cite{geiger2012automatic,yan2023joint} have also been directed towards reducing human intervention and making the calibration process more automated. There are also similar implementations \cite{chen2019omnidirectional, kholodilin2020omnidirectional} also adopted chessboards as calibration targets, tailoring their approaches for different applications. 

Target-based methods can achieve high-precision calibration but the major limitations resulting from the reliance on targets are unavoidable. The trade-offs between accuracy, accessibility and automation should always be considered and striking a good balance among these factors presents a great challenge.

\subsection{Target-Less Calibration}
To eliminate the requirement for calibration targets, target-less methods turn to utilize common features and avoid finding explicit correspondences. Edges \cite{moghadam2013line,levinson2013automatic,castorena2016autocalibration,bai2020lidar, zhu2021camvox,yuan2021pixel} and lane markings \cite{ma2021crlf,jeong2019road} are the most common features used in target-less calibration. After extracting features from two modalities, the extrinsics are optimized to maximize the consistency of the extracted features.

To avoid relying on predefined features, some methods \cite{nagy2019sfm,nagy2020fly} opt to recover 3D geometry through Structure-from-Motion (SfM) and jointly optimize extrinsic parameters in reconstruction. On the other hand, monocular depth estimation techniques have also been employed to convert the calibration problem into an intra-modality matching problem \cite{borer2024chaos}. Mutual information techniques \cite{pandey2012automatic,taylor2013automatic} are also utilized to evaluate the similarity between projected depth or intensity maps and RGB images and extrinsics are estimated by maximizing the mutual information.

Another category of target-less calibration methods \cite{taylor2016motion,ishikawa2018lidar} leverages motion information by matching trajectories or utilizes semantic information through instance matching. Motion-based methods require sufficient data from various motion patterns to ensure accuracy. Semantic-based methods depend on pretrained semantic segmentation models and require scenes containing multiple distinguishable instances.

Target-less methods introduce different features and metrics to formulate an optimization problem and then solve for the extrinsics, but the effectiveness of these methods is heavily influenced by the initial conditions, the conformity of assumptions, and the suitability of the scene.

\subsection{Learning-Based Calibration}
Existing learning-based methods utilize deep neural networks either to regress the parameters directly or to obtain correspondences that can subsequently be employed to solve the PnP problem. These approaches can be broadly categorized as regression-based and matching-based methods.

Matching-based methods replace manual feature extraction and matching with networks while preserving explicit geometric solving. Although these methods often require more computational effort due to sparse or dense matching, they exhibit better generalizability and interpretability.

CFNet \cite{lv2021cfnet} first introduced the concept of calibration flow and applied the EPnP algorithm \cite{lepetit2009ep} to estimate extrinsic parameters. DXQ-Net \cite{jing2022dxq} utilizes a differentiable pose estimation module and predicts uncertainty measures to filter reliable matches. A series of works \cite{sorrenti2020cmrnet++,cattaneo2024cmrnext,petek2024automatic} advance to demonstrate strong generalization capabilities in real-world settings for both localization and calibration. Other types of methods aim to realize sparse keypoint matching \cite{ye2021keypoint} or object instance matching \cite{sun2022atop} to reduce computational demands and improve generalizability.

RegNet \cite{schneider2017regnet} pioneered a learning-based calibration framework. It established a typical pipeline consisting of feature extraction, feature matching, and global regression, setting a precedent for subsequent regression-based methods. Subsequent works \cite{iyer2018calibnet,yuan2020rggnet,duan2023robust} introduce additional geometric constraints and propose loss functions that consider appearance consistency, geometric consistency, and Riemannian metrics. Various other approaches \cite{wu2021netcalib,zhao2021calibdnn,wu2022psnet,wang2023mrcnet} leverage commonly used deep learning modules to modify network architectures and achieve superior performance. Further improvements \cite{shi2020calibrcnn,nguyen2022calibbd,shang2022calnet} involve incorporating temporal losses to refine calibration. Additional modality adaptation modules \cite{zhu2022robust} are also employed to bridge domain gaps via stereo matching \cite{wu2021netcalib}, monocular depth estimation \cite{zhu2023calibdepth,zhu2022robust}, and edge extraction \cite{hu2023dedgenet}. Feature matching structures, such as the cost-volume layer \cite{lv2021lccnet,duan2023robust} or cross-attention mechanisms are also utilized to facilitate explicit matching. Additionally, there have been efforts to apply point cloud \cite{wang2022fusionnet} and graph processing techniques \cite{zhu2023robust} to address the calibration problem.

While matching-based methods are advancing to generalize to unseen environments and sensor setups, regression-based methods still remain constrained to re-training or fine-tuning. In this paper, we comprehensively review previous methods and explain the inherent limitations of regression-based approaches. 

\vspace{-0.2cm}
\section{Analysis}
In this section, we first state the problem formulation and then review the typical frameworks of regression-based and matching-based methods. Following this, we analyze the working principles of regression-based methods and design a series of experiments to confirm our viewpoint. Furthermore, we also revisit the widely used training dataset generation pipeline and describe its latent flaws. Based on our prior findings, we investigate the details of data generation and determine how these operations affect the final performance.

\subsection{Problem Formulation}
The primary goal of LiDAR-Camera calibration is to estimate a transformation matrix $\mathbf{T} \in \mathbb{R}^{4 \times 4}$ which consists of rotation matrix $\mathbf{R} \in \mathbb{R}^{3 \times 3} $ and $\mathbf{t} \in \mathbb{R}^{3}$ to ensure the data from the two sensors can be accurately aligned. Given an image $\mathbf{I} \in \mathbb{R}^{H\times W\times 3}$ and a point cloud $\mathbf{P} \in \mathbb{R}^{N\times 3}$, the transformation between LiDAR 3D points and a camera 2D pixels can be represented by following equation:
\begin{equation}
\mathbf{p} = \mathbf{K} \mathbf{T} \mathbf{P}
= \mathbf{K} 
\begin{bmatrix}
\mathbf{R} & \mathbf{t}
\end{bmatrix}
\mathbf{P}
\end{equation}
where $\mathbf{p} \in \mathbb{R}^{N \times 2}$ denotes the locations of projected points on image plane and $\mathbf{K} \in \mathbb{R}^{3 \times 3}$ denotes the intrinsic parameters.

To be more specific, each 3D point $\begin{bmatrix} X_i & Y_i & Z_i \end{bmatrix}^{\top} \in \mathbb{R}^3$ is projected onto the image plane to get corresponding 2D coordinates $\begin{bmatrix} u_i & v_i \end{bmatrix}^{\top} \in \mathbb{R}^2$ and depth value $d_i$.
\begin{equation}
d_{i}
\begin{bmatrix}
u_{i}\\ 
v_{i}\\ 
1
\end{bmatrix}
= 
\begin{bmatrix}
f_x & 0 & c_x \\ 
0 & f_y & c_y \\ 
0 & 0 & 1 
\end{bmatrix}
\begin{bmatrix}
\mathbf{R} & \mathbf{t} \\
0 & 1 
\end{bmatrix}
\begin{bmatrix}
X_i \\ 
Y_i \\
Z_i \\
1
\end{bmatrix}
\end{equation}
where $f_{x}$ and $f_{y}$ denote the focal length along image width and height directions and $(c_x, c_y)$ denotes the optical center. We omit the conversion from homogeneous coordinates to non-homogeneous coordinates for the simplicity.

The classical target-based methods concentrate on finding the 2D-3D correspondences beforehand, then estimate the optimal extrinsic parameters $\mathbf{T}$ by minimizing the projection error of the matched 3D points and 2D pixels, which is specified by the following equation.
\begin{equation}
\hat{\mathbf{T}} = \text{argmin}\sum_{i=1}^{n}{\Vert \mathbf{K} {\mathbf{T}} \mathbf{P}_i - \hat{\mathbf{p}}_i \Vert}^{2}_2
\end{equation}

Current matching-based methods adopt a similar pattern but replace manual feature extraction and feature matching procedures with deep neural networks as shown in Fig. \ref{fig2} (b). While previous target-based methods can only implement sparse keypoint matching, matching-based methods can achieve semi-dense or dense matching with the help of feature matching and flow estimation techniques.

For current regression-based methods, they predict the extrinsic directly in an end-to-end manner as shown in Fig. \ref{fig2} (a). The typical framework of these methods consists of a dual-branch feature extractor, a feature fusion module and a global regressor. The framework abandons the explicit matching operation and relies on neural networks to address all issues. The initial training objective can be formulated as follows:
\begin{figure}[htbp]
\vspace*{0.2cm}
\centering
\includegraphics[width=8cm]{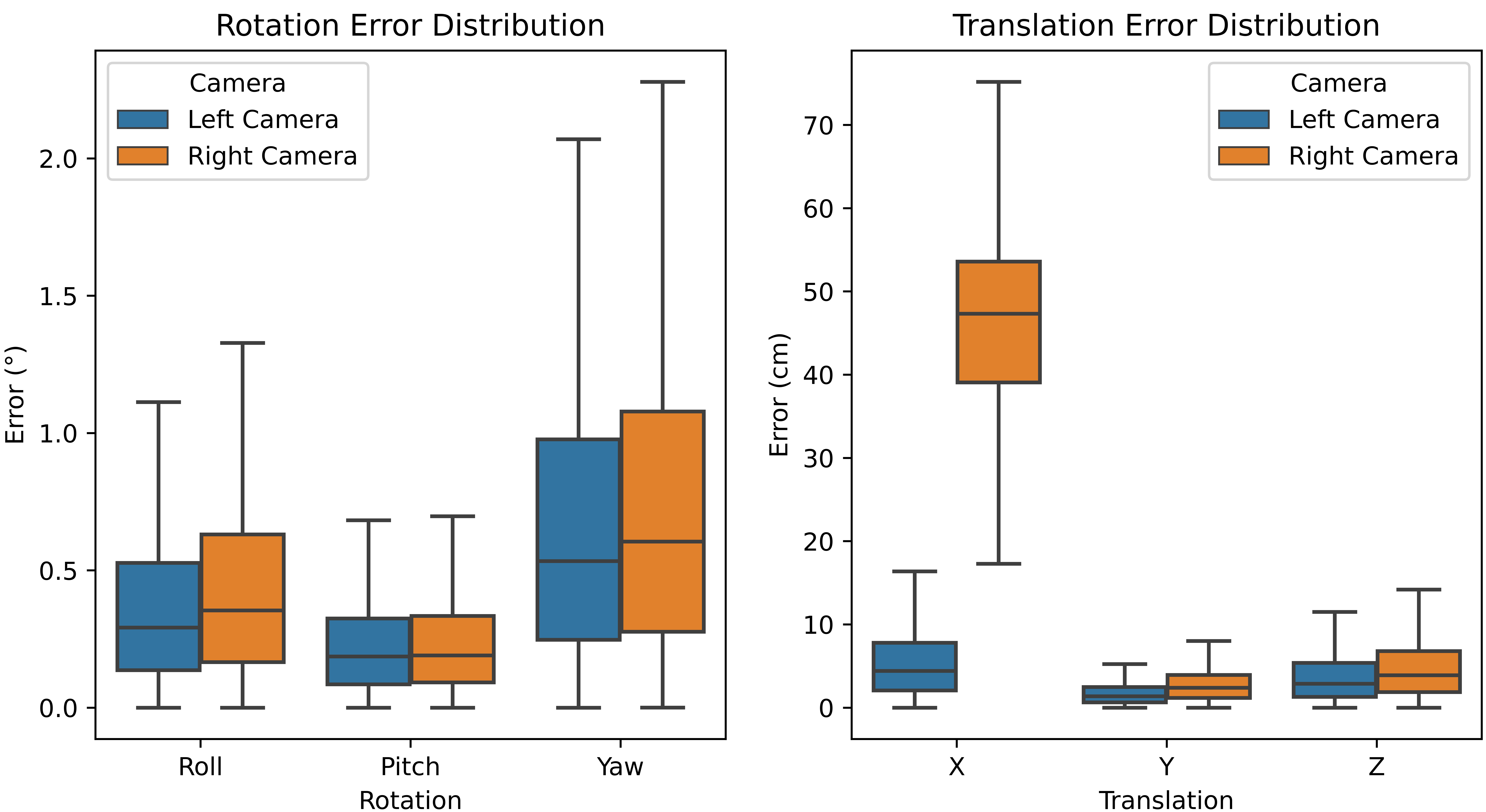}
\vspace*{-0.2cm}
\caption{Rotation and translation error distributions of the cross-camera test, which are derived from the single-branch test (\SI{\pm0.5}{\meter} / \SI{\pm5}{\degree}) with only depth as input, show minor variations for most components, except for the x-axis translation.}
\label{fig3}
\end{figure}

\begin{figure}[htbp]
\centering
\includegraphics[width=8cm]{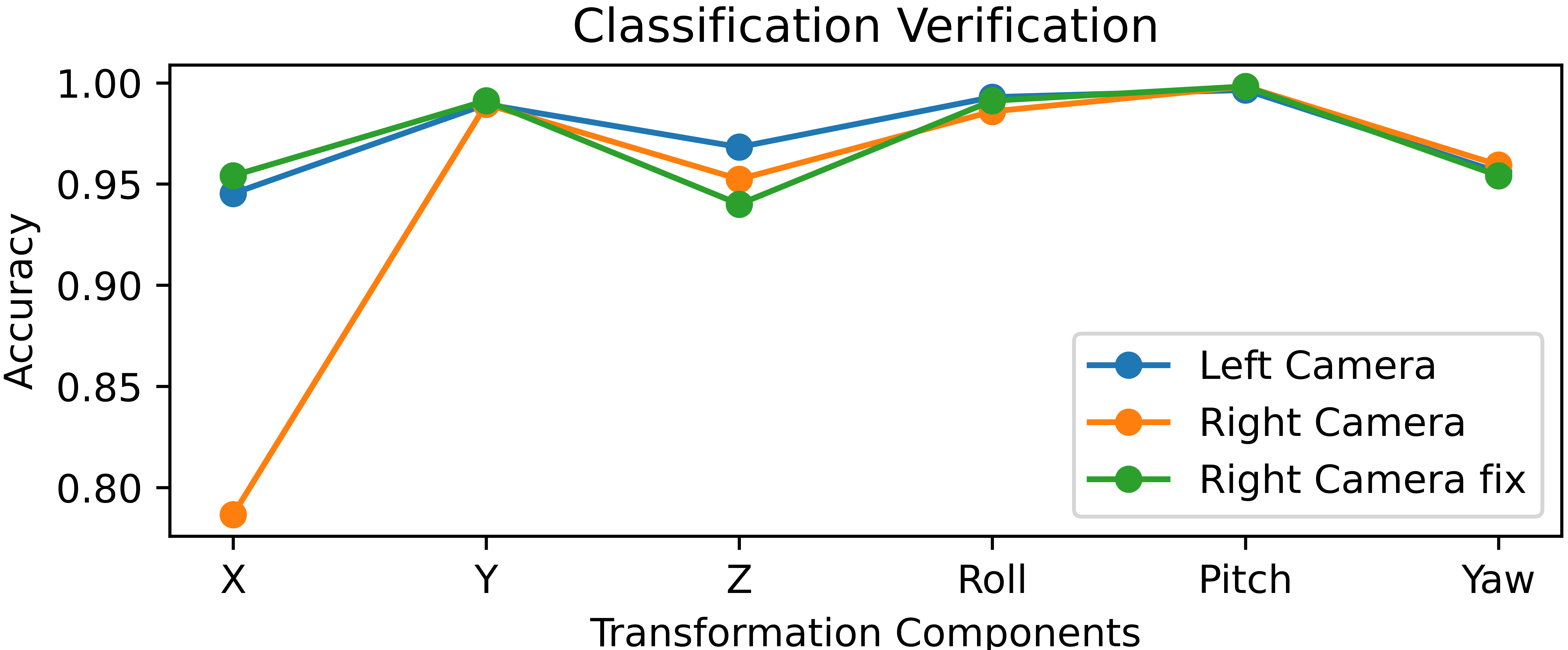}
\vspace*{-0.2cm}
\caption{The classification accuracy comparisons of the cross-camera test, which are derived from the dual-branch test (\SI{\pm1.5}{\meter} / \SI{\pm20}{\degree}), show a significant performance drop in x-axis translation. After regenerating labels using ground truth matrices multiplied by the transformation from the left camera to the right camera, the accuracy can be restored to the previous level.} 
\label{fig4}
\vspace*{-0.5cm}
\end{figure}

\begin{equation}
\hat{\mathbb{F}} = \mathop{\text{argmin}}\limits_{\mathbb{F}} {\Vert \mathbb{F}{(\mathbf{P} \text{,} \mathbf{I}) - \mathbf{T}_{gt}} \Vert}
\end{equation}
where $\mathbb{F}$ represents the neural network which takes point cloud and image as input and predicts the extrinsic parameters. Considering the architectural differences for extracting point cloud and image features, most works choose to project the point cloud to a depth map then utilize symmetrical networks to extract features. After using depth map as an intermediate representation, the training objective changes as follows:
\begin{equation}
\hat{\mathbb{F}} = \mathop{\text{argmin}}\limits_{\mathbb{F}} {\Vert \mathbb{F}{(\mathbf{D} \text{,} \mathbf{I}) - \mathbf{T}_{gt}} \Vert}
\end{equation}
where $\mathbf{D}$ denotes the projected depth map.

The two formulations seem equivalent, but the projection process explicitly involves intrinsic parameters . Starting from basic geometric principles, it is clear that calibration should take intrinsic parameters into account. Most classical  calibration methods acquire initial intrinsic parameters to reduce the complexity of the problem and improve the stability of the solution. However, almost all regression-based methods either ignore the intrinsic parameters or only involve them implicitly. On the other hand, the framework of regression-based methods is similar to that of image classification, but image classification focuses more on high-level semantic extraction. For LiDAR-Camera calibration, the matching relationship and positional information are more critical. This raises two important questions:
\begin{itemize}
\item Do neural networks need to take intrinsic parameters as input or not?
\item If intrinsic information is necessary but omitted in most implementations, what have neural networks learned?
\end{itemize}

\begin{figure}[htbp]
\centering
\includegraphics[width=8cm]{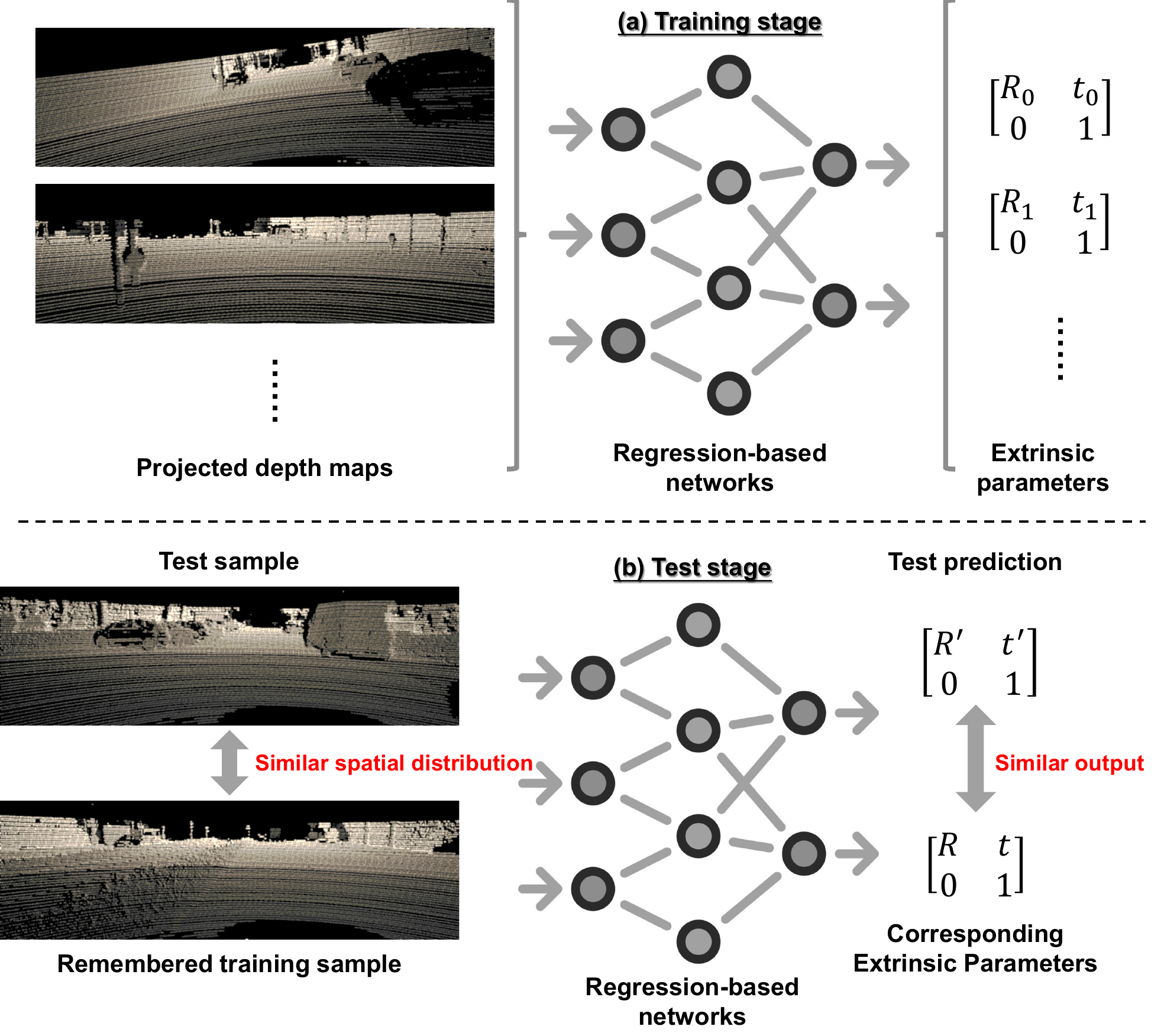}
\vspace*{-0.2cm}
\caption{The working principles of regression-based methods are illustrated. In the training stage, the network learns to memorize the mapping relationship between depth map distributions and extrinsic parameters. In the test stage, the network predicts the extrinsics based on the similarity between training data and test samples.}
\label{fig5}
\vspace*{-0.5cm}
\end{figure}

\subsection{Working Principles}
In order to uncover the working principles of regression-based methods, we establish a simple framework composed of only basic modules. We observe that the network modifications primarily concentrated on introducing extra modality adaption, multi-level feature extraction and feature matching modules. Although related works claim that their modifications are reasonable and effective, they do not alter the overall architecture.

Specifically, we utilize the ResNet-18 \cite{he2016deep} backbone to build two symmetrical encoders for extracting RGB and depth features, following the implementation of CalibNet \cite{iyer2018calibnet}. After that, the extracted features from the two branches are concatenated together and feed into fully connected layers to regress rotation and translation parameters. We perform experiments on the KITTI Odometry \cite{Geiger2012CVPR} dataset and the KITTI-360 \cite{liao2022kitti} dataset. Sequences 01 to 20 in the KITTI Odometry dataset are utilized for training and validation. Sequences 00 in the KITTI Odometry dataset and SLAM test sequences in the KITTI-360 dataset are utilized for test. For training the networks, we only adopt L2 loss , and Adam Optimizer \cite{kingma2014adam} with initial learning rate $1e^{-4}$.

\begin{table*}
\vspace{0.2cm}
\scriptsize
\centering
\caption{Calibration Error on the KITTI Dataset}
\vspace{-0.3cm}
\label{tab:1}
\setlength\tabcolsep{4.0pt}
\begin{threeparttable}
\resizebox{\textwidth}{!}
{
    \begin{tabular}{l c | cc !{\color{gray}\vline} ccc ccc | cc !{\color{gray}\vline} ccc ccc}
        \toprule
        & & \multicolumn{8}{c|}{\textbf{Left camera}} & \multicolumn{8}{c}{\textbf{Right camera}} \\
        & & \multicolumn{2}{c}{Magnitude} & \multicolumn{3}{c}{Translation [cm]} & \multicolumn{3}{c|}{Rotation [\degree]} & \multicolumn{2}{c}{Magnitude} & \multicolumn{3}{c}{Translation [cm]} & \multicolumn{3}{c}{Rotation [\degree]} \\
        \textbf{Method} & \textbf{Initial range} & $E_\textrm{t}$ [cm] & $E_\textrm{R}$ [\degree] & x & y & z & roll & pitch & yaw & $E_\textrm{t}$ [cm] & $E_\textrm{R}$ [\degree] & x & y & z & roll & pitch & yaw \\
        \midrule
        CMRNet \cite{cattaneo2019cmrnet} & \SI{\pm2}{\meter} / \SI{\pm10}{\degree} & 26.26 & 1.09 & 16.80 & 4.20 & 15.23 & 0.62 & 0.26 & 0.66 & 61.00 & 1.10 & 56.20 & 4.20 & 15.33 & 0.63 & 0.26 & 0.66 \\
        RGGNet \cite{yuan2020rggnet} & \SI{\pm0.3}{\meter} / \SI{\pm20}{\degree} & 11.49 & 1.29 & 8.14 & 2.79 & 3.97 & 0.64 & 0.35 & 0.74 & 23.52 & 3.87 & 18.03 & 5.55 & 6.06 & 1.48 & 0.51 & 3.38 \\
        LCCNet \cite{lv2021lccnet} & \SI{\pm1.5}{\meter} / \SI{\pm20}{\degree} & 1.31 & 0.19 & 0.24 & 0.48 & 1.11 & 0.03 & 0.02 & 0.02 & 52.48 & 1.56 & 52.42 & 0.48 & 1.30 & 0.04 & 0.02 & 1.56 \\
        CMRNext \cite{cattaneo2024cmrnext} & \SI{\pm1.5}{\meter} / \SI{\pm20}{\degree} & 1.89 & 0.08 & 1.12 & 0.83 & 0.79 & 0.04 & 0.04 & 0.04 & 7.07 & 0.23 & 2.17 & 5.78 & 0.94 & 0.20 & 0.05 & 0.05\\
        MDPCalib \cite{petek2024automatic} & -- & 0.18 & 0.06 & 0.07 & 0.16 & 0.01 & 0.04 & 0.02 & 0.04 & 2.94 & 0.14 & 0.66 & 2.78 & 0.49 & 0.13 & 0.03 & 0.05 \\
        \grayrule
        Dual-branch + RGB + Depth &\SI{\pm1.5}{\meter} / \SI{\pm20}{\degree} & 37.52 & 4.50 & 30.71 & 18.24 & 26.11 & 3.17 & 2.05 & 3.46 & 50.33 & 5.91 & 30.70 & 18.24 & 26.11 & 3.17 & 2.05 & 3.46 \\
        Dual-branch + RGB + Depth &\SI{\pm1.0}{\meter} / \SI{\pm10}{\degree} & 25.00 & 2.53 & 17.33 & 6.21 & 12.37 & 1.65 & 0.67 & 1.32 & 47.58 & 2.69 & 42.15 & 6.86 & 13.74 & 1.78 & 0.68 & 1.40 \\
        Dual-branch + RGB + Depth &\SI{\pm0.5}{\meter} / \SI{\pm5}{\degree} & 12.02 & 1.37 & 8.96 & 2.60 & 5.47 & 0.67 & 0.33 & 0.95 & 43.39 & 1.55 & 41.98 & 3.21 & 6.68 & 0.84 & 0.34 & 1.03 \\
        Single-branch + RGB + Depth & \SI{\pm0.5}{\meter} / \SI{\pm5}{\degree} & 12.19 & 1.27 & 8.73 & 2.97 & 5.83 & 0.60 & 0.30 & 0.89 & 45.28 & 1.39 & 43.77 & 3.49 & 7.04 & 0.68 & 0.30 & 0.98 \\
        Single-branch + Depth &\SI{\pm0.5}{\meter} / \SI{\pm5}{\degree} & 8.03 & 0.96 & 5.71 & 1.80 & 3.97 & 0.38 & 0.23 & 0.72 & 46.10 & 1.05 & 45.40 & 2.80 & 4.96 & 0.46 & 0.24 & 0.78 \\
        Single-branch + Depth $^\dagger$ &\SI{\pm0.5}{\meter} / \SI{\pm5}{\degree} & 18.18 & 1.11 & 16.58 & 2.13 & 4.30 & 0.48 & 0.25 & 0.82 & 20.99 & 1.18 & 19.42 & 2.16 & 4.48 & 0.51 & 0.26 & 0.88 \\
        \bottomrule
    \end{tabular}
}
\end{threeparttable}
\footnotesize
\begin{flushleft}
Regression-based methods, including CMRNet, RGGNet, and LCCNet, fail to generalize to different cameras. Matching-based methods such as CMRNet and MDPCalib show significantly better generalizability.
We perform dual-branch tests on different initial ranges to verify that the phenomenon consistently exists.
$^\dagger$: This result is derived from mixed training utilizing left camera-LiDAR and right camera-LiDAR pairs. 
The performance drops for both cameras, indicating that network is diffused by the different parameters mapping.
\end{flushleft}
\vspace*{-0.8cm}
\end{table*}

To validate whether the matching relationships determine the predictions, we train our network using only the left camera images and test the calibration accuracy using both left and right camera images. Since the two cameras are of the same type and are oriented in the same direction, there is only a slight difference in their intrinsic parameters and fields of view. If neural networks infer extrinsics according to the matching relationships, the accuracy of calibrating LiDAR to the right camera should be comparable. However, we observe that the trained networks suffer from a significant performance drop in translation but exhibit relatively slight performance degradation in rotation as shown in Tab. \ref{tab:1}. We also perform experiments on different approaches and observe the similar phenomenon. The error distributions of the simple generalizability test are illustrated in Fig. \ref{fig3}.

The performance degradation in rotation prediction is relatively acceptable since only left camera data is incorporated in the training data, and the trained models may be overfit to the left camera images. Nevertheless, the performance drop in translation prediction is significant and is primarily caused by horizontal errors. It is worth noting that the translation error is numerically close to the camera baseline. A reasonable explanation for this phenomenon is that neural networks actually learn to map the spatial distributions of the depth map to corresponding parameters. We hypothesize that the trained models are not primarily overfitting to the left camera images but to the transformation between the LiDAR and the left camera.

To further verify our viewpoint, we remove the RGB feature extractor and retain a single-branch feature extractor that takes a 4-channel RGB-D image or only a depth map as input. Referring to Tab. \ref{tab:1}, the calibration accuracy of this single-branch architecture is comparable to that of the dual-branch architecture. The performance difference between using RGB-D images and depth maps as inputs is also trivial. This suggests that the proposed networks focus more on the spatial distributions of the depth map rather than the matching relationships. Given that the spatial relationship between the camera and LiDAR is fixed, the training data is generated by simulating random perturbations. Intuitively, similar perturbations produce similar spatial distributions in the depth maps, regardless of the changes in the scenes perceived by the sensors. Conversely, if the spatial distributions are similar, the predicted values should also be similar. As shown in Fig. \ref{fig5}, the fundamental working principle of regression-based methods appears to be clustering relevant depth map patterns and memorizing the mappings. The actual training target can be simplified to
\begin{equation}
\hat{\mathbb{F}} = \mathop{\text{argmin}}\limits_{\mathbb{F}} {\Vert \mathbb{F}{(\mathbf{D}) - \mathbf{T}_{gt}} \Vert}
\end{equation}
which means these models do not operate according to the expectations claimed in previous works.

Although several methods like LCCNet \cite{lv2021lccnet} introduce designed layers for explicit matching, the trained models still yield similar results to our simple framework without extra modules. That indicates that performance improvements are mainly attributed to additional module complexity rather than their specific design.

To validate our argument from an alternative viewpoint, we modify the global regression layer to a classification layer that determines whether each transformation component is greater than or less than the ground truth value. After converting the problem to a classification task, the classification accuracy still drops severely in the x-axis translation. However, if we regenerate the labels by considering the relative transformation, the accuracy can recover to a comparable status. The accuracy comparisons are shown in Fig. \ref{fig4}.

Both regression and classification experiments verify that the networks only learn to remember the mapping relationship within the $\mathbf{SE(3)}$ space. However, this mapping relationship is derived from and constrained by the ground truth extrinsics and random perturbations. If the ground truth extrinsics change when applied to another LiDAR-Camera pair, the learned mapping relationship is no longer valid.

\subsection{Limitations of Dataset Generation}

Collecting high-quality datasets is one of the most important prerequisites in training neural networks. However, it is not trivial to collect a diverse LiDAR-Camera calibration dataset. The main reason is that the calibration process is complex and time-consuming, making it impractical to create numerous sensor layouts and obtain accurate calibration results in reality. For most datasets, the sensor layouts are fixed, and the provided extrinsic parameters remain stable over different time periods. To address the issue of limited data availability, RegNet\cite{schneider2017regnet} first proposed to generate a large amount of training data by varying the perturbations. Given an initial transformation $\mathbf{T}_{init} \in \mathbb{R}^{4 \times 4}$ and random perturbation $\Delta \mathbf{T} \in \mathbb{R}^{4 \times 4}$, the corresponding ground truth extrinsic should be:
\begin{equation}
\mathbf{T}_{gt} = \mathbf{T}_{init} (\Delta T)^{-1}
\end{equation}

\begin{figure}[htbp]
\vspace*{0.2cm}
\centering
\includegraphics[width=8cm, height=4cm]{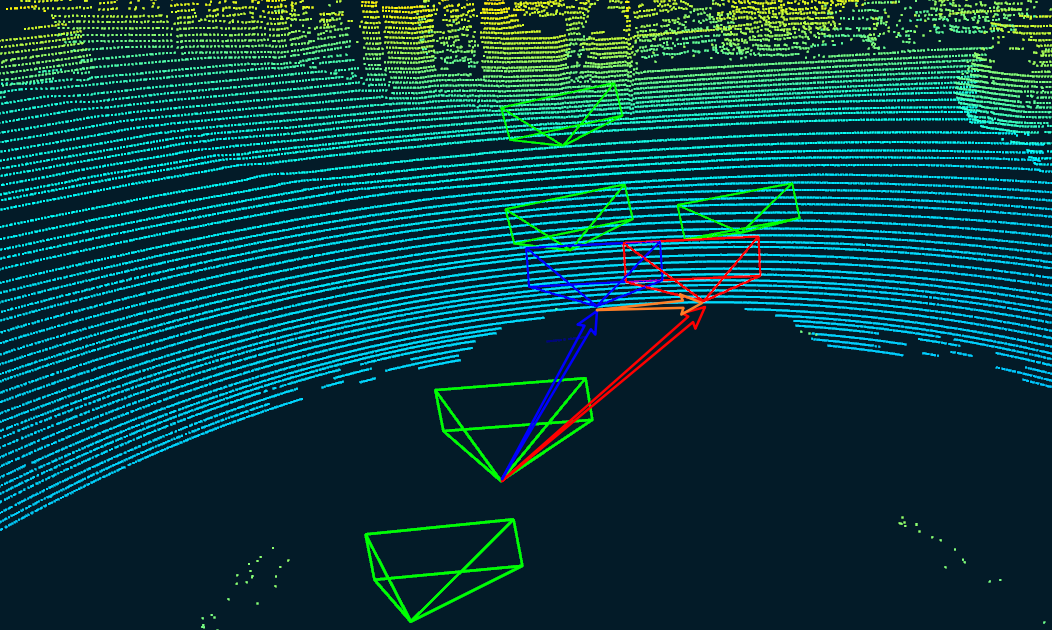}
\vspace*{-0.3cm}
\caption{Illustration of the performance drop in cross-camera generalization tests. The red frustum and blue frustum represent left camera and right camera, respectively. The green frusta represent the perturbed virtual viewpoints. Since the networks only learn the mapping relationship from depth maps to perturbations based on left camera's extrinsics, there is a bias in the prediction results. An extra transformation from left camera to right camera is required to correct this bias.}
\label{fig6}
\vspace*{-0.5cm}
\end{figure}

This generation pipeline seems reasonable and has been consistently adopted by most previous methods. Despite this, there are two important factors that have been neglected up until now. The first factor is that such random perturbations cannot simulate actual variations in sensor configurations, especially when the perturbation values are substantial. In practice, point clouds captured by a LiDAR change with different placement positions, and the overlapping regions of the fields of view also vary. Simply transforming point clouds by perturbations does not produce realistic data. The second factor is that random perturbations generated by uniform sampling lead networks to overfit to the current sensor setup, preventing them from generalizing to other unseen configurations, as previously stated and illustrated in Fig. \ref{fig6}. This method only generates numerous de-calibration data rather than varying extrinsic calibration data. Therefore, we argue that this generation pipeline cannot create ideal datasets and leads networks to fall into an overfitting trap. 

At the same time, it is common to assume that intrinsic parameters can be attained in advance and the main target is to estimate the extrinsic parameters. Nevertheless, the intrinsic parameters are only utilized in depth map projection and depth maps are always resized to other size. It is noteworthy that intrinsic parameters should change when operations like cropping, padding and resizing are performed.

We attempted to implement random cropping in network training, but observed a decrease in accuracy following such random augmentation. The results shown in Tab. \ref{tab:2} indicate that previous works rely on stable intrinsic parameters, and consistent preprocessing operations. While projection does involve intrinsic parameters in network training, it introduces an inductive bias. Random cropping causes the intrinsic parameters of each projected depth map to vary, thereby breaking the prior network bias. The generalization tests on KITTI-360 \cite{liao2022kitti} also verify the claim for the significant performance drop caused by intrinsic and extrinsic variations. 

\begin{table}
\scriptsize
\centering
\caption{Data Generation Limitation Analysis}
\vspace{-0.3cm}
\label{tab:2}
\setlength\tabcolsep{3.0pt}
\begin{threeparttable}
    \begin{tabular}{l | cc !{\color{gray}\vline} ccc ccc}
        \toprule
        & \multicolumn{2}{c}{Magnitude} & \multicolumn{3}{c}{Translation [cm]} & \multicolumn{3}{c}{Rotation [\degree]} \\
        \textbf{Component} & $E_\textrm{t}$ [cm] & $E_\textrm{R}$ [\degree] & x & y & z & roll & pitch & yaw \\
        \midrule
        Single-branch + Depth & 8.03 & 0.96 & 5.71 & 1.80 & 3.97 & 0.38 & 0.23 & 0.72 \\
        Random cropping: & & & \\
        \hspace{.5pt} + Vertical  & 11.07 & 2.35 & 6.46 & 2.06 & 7.13 & 0.49 & 1.93 & 0.84\\
        \hspace{.5pt} + Horizontal & 11.56 & 1.98 & 9.21 & 2.25 & 4.56 & 0.63 & 0.28 & 1.71\\
        \hspace{.5pt} + Both  & 13.53 & 2.97 & 9.06 & 2.34 & 7.73 & 0.68 & 1.90 & 1.79\\
        \grayrule
        Generalization test:  & & & \\
        \hspace{.5pt} + KITTI-360 cam0  & 41.57 & 3.67 & 28.81 & 22.66 & 11.55 & 1.05 & 0.90 & 3.10\\
        \hspace{.5pt} + KITTI-360 cam1  & 76.74 & 3.56 & 70.34 & 22.96 & 12.88 & 1.34 & 0.94 & 2.79\\
        \bottomrule
    \end{tabular}
    \footnotesize
\end{threeparttable}
\vspace{-0.7cm}
\end{table}

\subsection{Regression Clues}
As the depth map distribution is the primary focus of regression-based methods, we investigate the main regression clues by validating the performance of pretrained models with different operations. From the experimental results in Table \ref{tab:3}, it can be observed that vertical and horizontal perturbations can influence network predictions. The results align with the intuition that pitch errors cause changes in the vertical direction, while x-axis translation and yaw errors cause changes in the horizontal direction.

We also introduce densification and sampling to simulate different point cloud densities, as the number of LiDAR beams and scanning patterns vary in applications. Our findings indicate that networks are sensitive to densification when the initial input is sparse, particularly for translation prediction. Random sampling also causes performance degradation, but the impact is acceptable when the sample ratio is high. As the sample ratio decreases, performance gradually declines.

The experimental results indicate that regression-based methods learn to exploit specific spatial locations and patterns for predictions. These networks tend to overfit to specific settings, including ground truth extrinsic parameters, intrinsic parameters, point cloud densities, and preprocessing operations. For practical applications, the main challenge is to eliminate reliance on certain clues by creating a canonical model or incorporating all relevant factors into the networks.

\begin{table}
\vspace{0.2cm}
\scriptsize
\centering
\caption{Regression Clues Analysis}
\vspace{-0.3cm}
\label{tab:3}
\setlength\tabcolsep{3.0pt}
\begin{threeparttable}
    \begin{tabular}{l | cc !{\color{gray}\vline} ccc ccc}
        \toprule
        & \multicolumn{2}{c}{Magnitude} & \multicolumn{3}{c}{Translation [cm]} & \multicolumn{3}{c}{Rotation [\degree]} \\
        \textbf{Component} & $E_\textrm{t}$ [cm] & $E_\textrm{R}$ [\degree] & x & y & z & roll & pitch & yaw \\
        \midrule
        Single-branch + Depth & 8.03 & 0.96 & 5.71 & 1.80 & 3.97 & 0.38 & 0.23 & 0.72 \\
        Vertical perturb.: & & & \\
        \hspace{.4cm} + 5 pixels & 9.25 & 1.06 & 5.97 & 2.08 & 5.29 & 0.43 & 0.33 & 0.75 \\
        \hspace{.4cm} + 10 pixels & 10.28 & 1.40 & 6.40 & 2.34 & 6.17 & 0.51 & 0.71 & 0.87 \\
        \hspace{.4cm} + 20 pixels & 12.46 & 2.01 & 7.28 & 2.73 & 8.01 & 0.61 & 1.42 & 0.92 \\
        \hspace{.4cm} + 40 pixels & 19.42 & 3.45 & 9.59 & 4.55 & 13.73 & 0.88 & 2.75 & 1.21 \\
        \grayrule
        Horizontal perturb.: & & & \\
        \hspace{.4cm} + 5 pixels & 7.96 & 0.94 & 5.66 & 1.70 & 3.90 & 0.38 & 0.23 & 0.70 \\
        \hspace{.4cm} + 10 pixels & 8.06 & 0.95 & 5.74 & 1.76 & 3.99 & 0.38 & 0.23 & 0.71 \\
        \hspace{.4cm} + 20 pixels & 9.49 & 1.47 & 6.79 & 1.96 & 4.78 & 0.41 & 0.24 & 1.31 \\
        \hspace{.4cm} + 40 pixels & 15.40 & 2.34 & 12.41 & 2.91 & 6.49 & 0.44 & 0.26 & 2.22 \\
        \hspace{.4cm} + 80 pixels & 29.67 & 4.03 & 27.42 & 5.36 & 7.23 & 0.59 & 0.32 & 3.91 \\
        \grayrule
        Densification: & & & \\
        \hspace{.4cm} 3$\times$3 kernel & 82.92 & 2.62 & 17.72 & 16.96 & 76.00 & 1.21 & 1.47 & 1.28 \\
        \hspace{.4cm} 5$\times$5 kernel & 98.49 & 3.78 & 23.53 & 27.57 & 87.15 & 2.03 & 2.12 & 1.63 \\
        \hspace{.4cm} 7$\times$7 kernel & 102.80 & 4.06 & 24.48 & 29.69 & 90.80 & 2.23 & 2.03 & 1.71 \\
        \grayrule
        Downsample: & & & \\
        \hspace{.4cm} sample ratio 0.9 & 9.90 & 1.05 & 6.97 & 2.22 & 5.09 & 0.44 & 0.29 & 0.76 \\
        \hspace{.4cm} sample ratio 0.8 & 17.30 & 1.42 & 10.49 & 4.39 & 10.59 & 0.64 & 0.58 & 0.89 \\
        \hspace{.4cm} sample ratio 0.5 & 43.99 & 2.24 & 22.16 & 21.28 & 23.43 & 1.36 & 0.81 & 1.17 \\
        \bottomrule
    \end{tabular}
    \footnotesize
    The input depth map size is $320 \times 1024$.
\end{threeparttable}
\vspace*{-0.7cm}
\end{table}

\section{Conclusion}
In this paper, we review previous learning-based LiDAR-Camera calibration methods and identify the limitations of regression-based methods and the data generation pipeline through theoretical analysis and a series of experiments. We reveal that regression-based methods bypass the claimed feature matching and focus more on the depth map distribution. The data generation pipeline does not provide suitable data for actually solving the calibration problem within the regression-based framework. Although these methods report impressive performance, there is still a long way to go for real applications. We invite the community to pay more attention to the underlying principles and believe that matching-based methods combined with more explicit geometric constraints are a promising solution.

\bibliographystyle{unsrt}
\bibliography{ref}

\begin{thebibliography}{10}

\bibitem{zheng2022fast}
Chunran Zheng, Qingyan Zhu, Wei Xu, Xiyuan Liu, Qizhi Guo, and Fu~Zhang.
\newblock Fast-livo: Fast and tightly-coupled sparse-direct
  lidar-inertial-visual odometry.
\newblock In {\em 2022 IEEE/RSJ international conference on intelligent robots
  and systems (IROS)}, pages 4003--4009. IEEE, 2022.

\bibitem{liu2023bevfusion}
Zhijian Liu, Haotian Tang, Alexander Amini, Xinyu Yang, Huizi Mao, Daniela~L
  Rus, and Song Han.
\newblock Bevfusion: Multi-task multi-sensor fusion with unified bird's-eye
  view representation.
\newblock In {\em 2023 IEEE international conference on robotics and automation
  (ICRA)}, pages 2774--2781. IEEE, 2023.

\bibitem{mateen2022smart}
Abdul Mateen, Muhammad~Zahid Hanif, Narayan Khatri, Sihyung Lee, and Seung~Yeob
  Nam.
\newblock Smart roads for autonomous accident detection and warnings.
\newblock {\em Sensors}, 22(6):2077, 2022.

\bibitem{bai2022transfusion}
Xuyang Bai, Zeyu Hu, Xinge Zhu, Qingqiu Huang, Yilun Chen, Hongbo Fu, and
  Chiew-Lan Tai.
\newblock Transfusion: Robust lidar-camera fusion for 3d object detection with
  transformers.
\newblock In {\em Proceedings of the IEEE/CVF conference on computer vision and
  pattern recognition}, pages 1090--1099, 2022.

\bibitem{liu2023uniseg}
Youquan Liu, Runnan Chen, Xin Li, Lingdong Kong, Yuchen Yang, Zhaoyang Xia,
  Yeqi Bai, Xinge Zhu, Yuexin Ma, Yikang Li, et~al.
\newblock Uniseg: A unified multi-modal lidar segmentation network and the
  openpcseg codebase.
\newblock In {\em Proceedings of the IEEE/CVF International Conference on
  Computer Vision}, pages 21662--21673, 2023.

\bibitem{wang2022rail}
Yusheng Wang, Weiwei Song, Yidong Lou, Yi~Zhang, Fei Huang, Zhiyong Tu, and
  Qiangsheng Liang.
\newblock Rail vehicle localization and mapping with lidar-vision-inertial-gnss
  fusion.
\newblock {\em IEEE Robotics and Automation Letters}, 7(4):9818--9825, 2022.

\bibitem{wang2023openoccupancy}
Xiaofeng Wang, Zheng Zhu, Wenbo Xu, Yunpeng Zhang, Yi~Wei, Xu~Chi, Yun Ye,
  Dalong Du, Jiwen Lu, and Xingang Wang.
\newblock Openoccupancy: A large scale benchmark for surrounding semantic
  occupancy perception.
\newblock In {\em Proceedings of the IEEE/CVF International Conference on
  Computer Vision}, pages 17850--17859, 2023.

\bibitem{zhang2000flexible}
Zhengyou Zhang.
\newblock A flexible new technique for camera calibration.
\newblock {\em IEEE Transactions on pattern analysis and machine intelligence},
  22(11):1330--1334, 2000.

\bibitem{moghadam2013line}
Peyman Moghadam, Michael Bosse, and Robert Zlot.
\newblock Line-based extrinsic calibration of range and image sensors.
\newblock In {\em 2013 IEEE International Conference on Robotics and
  Automation}, pages 3685--3691. IEEE, 2013.

\bibitem{levinson2013automatic}
Jesse Levinson and Sebastian Thrun.
\newblock Automatic online calibration of cameras and lasers.
\newblock In {\em Robotics: science and systems}, volume~2. Citeseer, 2013.

\bibitem{nagy2019sfm}
Bal{\'a}zs Nagy, Levente Kov{\'a}cs, and Csaba Benedek.
\newblock Sfm and semantic information based online targetless camera-lidar
  self-calibration.
\newblock In {\em 2019 IEEE International Conference on Image Processing
  (ICIP)}, pages 1317--1321. IEEE, 2019.

\bibitem{ma2021crlf}
Tao Ma, Zhizheng Liu, Guohang Yan, and Yikang Li.
\newblock Crlf: Automatic calibration and refinement based on line feature for
  lidar and camera in road scenes.
\newblock {\em arXiv preprint arXiv:2103.04558}, 2021.

\bibitem{schneider2017regnet}
Nick Schneider, Florian Piewak, Christoph Stiller, and Uwe Franke.
\newblock Regnet: Multimodal sensor registration using deep neural networks.
\newblock In {\em 2017 IEEE intelligent vehicles symposium (IV)}, pages
  1803--1810. IEEE, 2017.

\bibitem{iyer2018calibnet}
Ganesh Iyer, R~Karnik Ram, J~Krishna Murthy, and K~Madhava Krishna.
\newblock Calibnet: Geometrically supervised extrinsic calibration using 3d
  spatial transformer networks.
\newblock In {\em 2018 IEEE/RSJ International Conference on Intelligent Robots
  and Systems (IROS)}, pages 1110--1117. IEEE, 2018.

\bibitem{cattaneo2019cmrnet}
Daniele Cattaneo, Matteo Vaghi, Augusto~Luis Ballardini, Simone Fontana,
  Domenico~G Sorrenti, and Wolfram Burgard.
\newblock Cmrnet: Camera to lidar-map registration.
\newblock In {\em 2019 IEEE intelligent transportation systems conference
  (ITSC)}, pages 1283--1289. IEEE, 2019.

\bibitem{zhang2004extrinsic}
Qilong Zhang and Robert Pless.
\newblock Extrinsic calibration of a camera and laser range finder (improves
  camera calibration).
\newblock In {\em 2004 IEEE/RSJ International Conference on Intelligent Robots
  and Systems (IROS)(IEEE Cat. No. 04CH37566)}, volume~3, pages 2301--2306.
  IEEE, 2004.

\bibitem{unnikrishnan2005fast}
Ranjith Unnikrishnan and Martial Hebert.
\newblock Fast extrinsic calibration of a laser rangefinder to a camera.
\newblock {\em Robotics Institute, Pittsburgh, PA, Tech. Rep. CMU-RI-TR-05-09},
  2005.

\bibitem{pandey2010extrinsic}
Gaurav Pandey, James McBride, Silvio Savarese, and Ryan Eustice.
\newblock Extrinsic calibration of a 3d laser scanner and an omnidirectional
  camera.
\newblock {\em IFAC Proceedings Volumes}, 43(16):336--341, 2010.

\bibitem{yan2023joint}
Guohang Yan, Feiyu He, Chunlei Shi, Pengjin Wei, Xinyu Cai, and Yikang Li.
\newblock Joint camera intrinsic and lidar-camera extrinsic calibration.
\newblock In {\em 2023 IEEE International Conference on Robotics and Automation
  (ICRA)}, pages 11446--11452. IEEE, 2023.

\bibitem{liu2023robust}
Jiahui Liu, Xingqun Zhan, Cheng Chi, Xin Zhang, and Chuanrun Zhai.
\newblock Robust extrinsic self-calibration of camera and solid state lidar.
\newblock {\em Journal of Intelligent \& Robotic Systems}, 109(4):81, 2023.

\bibitem{debattisti2013automated}
Stefano Debattisti, Luca Mazzei, and Matteo Panciroli.
\newblock Automated extrinsic laser and camera inter-calibration using
  triangular targets.
\newblock In {\em 2013 IEEE Intelligent Vehicles Symposium (IV)}, pages
  696--701. IEEE, 2013.

\bibitem{bu2021calibration}
Zean Bu, Changku Sun, Peng Wang, and Hang Dong.
\newblock Calibration of camera and flash lidar system with a triangular
  pyramid target.
\newblock {\em Applied sciences}, 11(2):582, 2021.

\bibitem{liao2018extrinsic}
Qinghai Liao, Zhenyong Chen, Yang Liu, Zhe Wang, and Ming Liu.
\newblock Extrinsic calibration of lidar and camera with polygon.
\newblock In {\em 2018 IEEE International Conference on Robotics and
  Biomimetics (ROBIO)}, pages 200--205. IEEE, 2018.

\bibitem{zhang2024automatic}
Guanyu Zhang, Kunyang Wu, Jun Lin, Tianhao Wang, and Yang Liu.
\newblock Automatic extrinsic parameter calibration for camera-lidar fusion
  using spherical target.
\newblock {\em IEEE Robotics and Automation Letters}, 2024.

\bibitem{dhall2017lidar}
Ankit Dhall, Kunal Chelani, Vishnu Radhakrishnan, and K~Madhava Krishna.
\newblock Lidar-camera calibration using 3d-3d point correspondences.
\newblock {\em arXiv preprint arXiv:1705.09785}, 2017.

\bibitem{geiger2012automatic}
Andreas Geiger, Frank Moosmann, {\"O}mer Car, and Bernhard Schuster.
\newblock Automatic camera and range sensor calibration using a single shot.
\newblock In {\em 2012 IEEE international conference on robotics and
  automation}, pages 3936--3943. IEEE, 2012.

\bibitem{chen2019omnidirectional}
Xin Chen, Fuqiang Zhou, and Ting Xue.
\newblock Omnidirectional field of view structured light calibration method for
  catadioptric vision system.
\newblock {\em Measurement}, 148:106914, 2019.

\bibitem{kholodilin2020omnidirectional}
Ivan Kholodilin, Yuan Li, and Qinglin Wang.
\newblock Omnidirectional vision system with laser illumination in a flexible
  configuration and its calibration by one single snapshot.
\newblock {\em IEEE Transactions on Instrumentation and Measurement},
  69(11):9105--9118, 2020.

\bibitem{castorena2016autocalibration}
Juan Castorena, Ulugbek~S Kamilov, and Petros~T Boufounos.
\newblock Autocalibration of lidar and optical cameras via edge alignment.
\newblock In {\em 2016 IEEE International Conference on Acoustics, Speech and
  Signal Processing (ICASSP)}, pages 2862--2866. IEEE, 2016.

\bibitem{bai2020lidar}
Zixuan Bai, Guang Jiang, and Ailing Xu.
\newblock Lidar-camera calibration using line correspondences.
\newblock {\em Sensors}, 20(21):6319, 2020.

\bibitem{zhu2021camvox}
Yuewen Zhu, Chunran Zheng, Chongjian Yuan, Xu~Huang, and Xiaoping Hong.
\newblock Camvox: A low-cost and accurate lidar-assisted visual slam system.
\newblock In {\em 2021 IEEE International Conference on Robotics and Automation
  (ICRA)}, pages 5049--5055. IEEE, 2021.

\bibitem{yuan2021pixel}
Chongjian Yuan, Xiyuan Liu, Xiaoping Hong, and Fu~Zhang.
\newblock Pixel-level extrinsic self calibration of high resolution lidar and
  camera in targetless environments.
\newblock {\em IEEE Robotics and Automation Letters}, 6(4):7517--7524, 2021.

\bibitem{jeong2019road}
Jinyong Jeong, Younghun Cho, and Ayoung Kim.
\newblock The road is enough! extrinsic calibration of non-overlapping stereo
  camera and lidar using road information.
\newblock {\em IEEE Robotics and Automation Letters}, 4(3):2831--2838, 2019.

\bibitem{nagy2020fly}
Bal{\'a}zs Nagy and Csaba Benedek.
\newblock On-the-fly camera and lidar calibration.
\newblock {\em Remote Sensing}, 12(7):1137, 2020.

\bibitem{borer2024chaos}
Jack Borer, Jeremy Tschirner, Florian {\"O}lsner, and Stefan Milz.
\newblock From chaos to calibration: A geometric mutual information approach to
  target-free camera lidar extrinsic calibration.
\newblock In {\em Proceedings of the IEEE/CVF Winter Conference on Applications
  of Computer Vision}, pages 8409--8418, 2024.

\bibitem{pandey2012automatic}
Gaurav Pandey, James McBride, Silvio Savarese, and Ryan Eustice.
\newblock Automatic targetless extrinsic calibration of a 3d lidar and camera
  by maximizing mutual information.
\newblock In {\em Proceedings of the AAAI conference on artificial
  intelligence}, volume~26, pages 2053--2059, 2012.

\bibitem{taylor2013automatic}
Zachary Taylor and Juan Nieto.
\newblock Automatic calibration of lidar and camera images using normalized
  mutual information.
\newblock In {\em Robotics and Automation (ICRA), 2013 IEEE International
  Conference on}. Citeseer, 2013.

\bibitem{taylor2016motion}
Zachary Taylor and Juan Nieto.
\newblock Motion-based calibration of multimodal sensor extrinsics and timing
  offset estimation.
\newblock {\em IEEE Transactions on Robotics}, 32(5):1215--1229, 2016.

\bibitem{ishikawa2018lidar}
Ryoichi Ishikawa, Takeshi Oishi, and Katsushi Ikeuchi.
\newblock Lidar and camera calibration using motions estimated by sensor fusion
  odometry.
\newblock In {\em 2018 IEEE/RSJ International Conference on Intelligent Robots
  and Systems (IROS)}, pages 7342--7349. IEEE, 2018.

\bibitem{lv2021cfnet}
Xudong Lv, Shuo Wang, and Dong Ye.
\newblock Cfnet: Lidar-camera registration using calibration flow network.
\newblock {\em Sensors}, 21(23):8112, 2021.

\bibitem{lepetit2009ep}
Vincent Lepetit, Francesc Moreno-Noguer, and Pascal Fua.
\newblock Ep n p: An accurate o (n) solution to the p n p problem.
\newblock {\em International journal of computer vision}, 81:155--166, 2009.

\bibitem{jing2022dxq}
Xin Jing, Xiaqing Ding, Rong Xiong, Huanjun Deng, and Yue Wang.
\newblock Dxq-net: Differentiable lidar-camera extrinsic calibration using
  quality-aware flow.
\newblock In {\em 2022 IEEE/RSJ International Conference on Intelligent Robots
  and Systems (IROS)}, pages 6235--6241. IEEE, 2022.

\bibitem{sorrenti2020cmrnet++}
D~Sorrenti, C~Daniele, A~Valada, et~al.
\newblock Cmrnet++: Map and camera agnostic monocular visual localization in
  lidar maps.
\newblock In {\em Proceeding of ICRA 2020 Workshop on Emerging Learning and
  Algorithmic Methods for Data Association in Robotics https://sites. google.
  com/view/edat/home also available as arxiv report https://arxiv.
  org/abs/2004.13795}, 2020.

\bibitem{cattaneo2024cmrnext}
Daniele Cattaneo and Abhinav Valada.
\newblock Cmrnext: Camera to lidar matching in the wild for localization and
  extrinsic calibration.
\newblock {\em arXiv preprint arXiv:2402.00129}, 2024.

\bibitem{petek2024automatic}
K{\"u}rsat Petek, Niclas V{\"o}disch, Johannes Meyer, Daniele Cattaneo, Abhinav
  Valada, and Wolfram Burgard.
\newblock Automatic target-less camera-lidar calibration from motion and deep
  point correspondences.
\newblock {\em arXiv preprint arXiv:2404.17298}, 2024.

\bibitem{ye2021keypoint}
Chao Ye, Huihui Pan, and Huijun Gao.
\newblock Keypoint-based lidar-camera online calibration with robust geometric
  network.
\newblock {\em IEEE Transactions on Instrumentation and Measurement}, 71:1--11,
  2021.

\bibitem{sun2022atop}
Yi~Sun, Jian Li, Yuru Wang, Xin Xu, Xiaohui Yang, and Zhenping Sun.
\newblock Atop: An attention-to-optimization approach for automatic
  lidar-camera calibration via cross-modal object matching.
\newblock {\em IEEE Transactions on Intelligent Vehicles}, 8(1):696--708, 2022.

\bibitem{yuan2020rggnet}
Kaiwen Yuan, Zhenyu Guo, and Z~Jane Wang.
\newblock Rggnet: Tolerance aware lidar-camera online calibration with
  geometric deep learning and generative model.
\newblock {\em IEEE Robotics and Automation Letters}, 5(4):6956--6963, 2020.

\bibitem{duan2023robust}
Zaipeng Duan, Xuzhong Hu, Junfeng Ding, Pei An, Xiao Huang, and Jie Ma.
\newblock A robust lidar-camera self-calibration via rotation-based alignment
  and multi-level cost volume.
\newblock {\em IEEE Robotics and Automation Letters}, 9(1):627--634, 2023.

\bibitem{wu2021netcalib}
Shan Wu, Amnir Hadachi, Damien Vivet, and Yadu Prabhakar.
\newblock Netcalib: A novel approach for lidar-camera auto-calibration based on
  deep learning.
\newblock In {\em 2020 25th International Conference on Pattern Recognition
  (ICPR)}, pages 6648--6655. IEEE, 2021.

\bibitem{zhao2021calibdnn}
Ganning Zhao, Jiesi Hu, Suya You, and C-C~Jay Kuo.
\newblock Calibdnn: multimodal sensor calibration for perception using deep
  neural networks.
\newblock In {\em Signal Processing, Sensor/Information Fusion, and Target
  Recognition XXX}, volume 11756, pages 324--335. SPIE, 2021.

\bibitem{wu2022psnet}
Yi~Wu, Ming Zhu, and Ji~Liang.
\newblock Psnet: Lidar and camera registration using parallel subnetworks.
\newblock {\em IEEE Access}, 10:70553--70561, 2022.

\bibitem{wang2023mrcnet}
Hao Wang, Zhangyu Wang, Guizhen Yu, Songyue Yang, and Yang Yang.
\newblock Mrcnet: Multi-resolution lidar-camera calibration using optical
  center distance loss network.
\newblock {\em IEEE Sensors Journal}, 2023.

\bibitem{shi2020calibrcnn}
Jieying Shi, Ziheng Zhu, Jianhua Zhang, Ruyu Liu, Zhenhua Wang, Shengyong Chen,
  and Honghai Liu.
\newblock Calibrcnn: Calibrating camera and lidar by recurrent convolutional
  neural network and geometric constraints.
\newblock In {\em 2020 IEEE/RSJ International Conference on Intelligent Robots
  and Systems (IROS)}, pages 10197--10202. IEEE, 2020.

\bibitem{nguyen2022calibbd}
An~Duy Nguyen and Myungsik Yoo.
\newblock Calibbd: Extrinsic calibration of the lidar and camera using a
  bidirectional neural network.
\newblock {\em IEEE Access}, 10:121261--121271, 2022.

\bibitem{shang2022calnet}
Hongcheng Shang and Bin-Jie Hu.
\newblock Calnet: Lidar-camera online calibration with channel attention and
  liquid time-constant network.
\newblock In {\em 2022 26th International Conference on Pattern Recognition
  (ICPR)}, pages 5147--5154. IEEE, 2022.

\bibitem{zhu2022robust}
Angfan Zhu, Yang Xiao, Chengxin Liu, and Zhiguo Cao.
\newblock Robust lidar-camera alignment with modality adapted local-to-global
  representation.
\newblock {\em IEEE Transactions on Circuits and Systems for Video Technology},
  33(1):59--73, 2022.

\bibitem{zhu2023calibdepth}
Jiangtong Zhu, Jianru Xue, and Pu~Zhang.
\newblock Calibdepth: Unifying depth map representation for iterative
  lidar-camera online calibration.
\newblock In {\em 2023 IEEE International Conference on Robotics and Automation
  (ICRA)}, pages 726--733. IEEE, 2023.

\bibitem{hu2023dedgenet}
Yiyang Hu, Hui Ma, Leiping Jie, and Hui Zhang.
\newblock Dedgenet: Extrinsic calibration of camera and lidar with
  depth-discontinuous edges.
\newblock In {\em 2023 IEEE International Conference on Robotics and Automation
  (ICRA)}, pages 11439--11445. IEEE, 2023.

\bibitem{lv2021lccnet}
Xudong Lv, Boya Wang, Ziwen Dou, Dong Ye, and Shuo Wang.
\newblock Lccnet: Lidar and camera self-calibration using cost volume network.
\newblock In {\em Proceedings of the IEEE/CVF Conference on Computer Vision and
  Pattern Recognition}, pages 2894--2901, 2021.

\bibitem{wang2022fusionnet}
Guangming Wang, Jiahao Qiu, Yanfeng Guo, and Hesheng Wang.
\newblock Fusionnet: Coarse-to-fine extrinsic calibration network of lidar and
  camera with hierarchical point-pixel fusion.
\newblock In {\em 2022 International Conference on Robotics and Automation
  (ICRA)}, pages 8964--8970. IEEE, 2022.

\bibitem{zhu2023robust}
Jianxiao Zhu, Xu~Li, Qimin Xu, and Zhengliang Sun.
\newblock Robust online calibration of lidar and camera based on cross-modal
  graph neural network.
\newblock {\em IEEE Transactions on Instrumentation and Measurement}, 2023.

\bibitem{he2016deep}
Kaiming He, Xiangyu Zhang, Shaoqing Ren, and Jian Sun.
\newblock Deep residual learning for image recognition.
\newblock In {\em Proceedings of the IEEE conference on computer vision and
  pattern recognition}, pages 770--778, 2016.

\bibitem{Geiger2012CVPR}
Andreas Geiger, Philip Lenz, and Raquel Urtasun.
\newblock Are we ready for autonomous driving? the kitti vision benchmark
  suite.
\newblock In {\em Conference on Computer Vision and Pattern Recognition
  (CVPR)}, 2012.

\bibitem{liao2022kitti}
Yiyi Liao, Jun Xie, and Andreas Geiger.
\newblock Kitti-360: A novel dataset and benchmarks for urban scene
  understanding in 2d and 3d.
\newblock {\em IEEE Transactions on Pattern Analysis and Machine Intelligence},
  45(3):3292--3310, 2022.

\bibitem{kingma2014adam}
Diederik~P Kingma.
\newblock Adam: A method for stochastic optimization.
\newblock {\em arXiv preprint arXiv:1412.6980}, 2014.

\end{thebibliography}
\end{document}